\documentclass[letterpaper, 10 pt, conference]{ieeeconf}

\IEEEoverridecommandlockouts

\input{paper.sty}

\usepackage{graphics} 
\usepackage{epsfig} 
\usepackage{times} 
\usepackage{comment}

\title{\LARGE \bf Learning Provably Robust Motion Planners Using Funnel Libraries}

\author{Ali Ekin Gurgen, Anirudha Majumdar, and Sushant Veer
\thanks{Ali Ekin Gurgen, Anirudha Majumdar, and Sushant Veer are with the Department of Mechanical and Aerospace Engineering, Princeton University, Princeton, NJ, 08544. 
        Emails: {\tt\small \{agurgen, ani.majumdar, sveer\}@princeton.edu}.}%
\thanks{The authors were supported by the Office of Naval Research [N00014-21-1-2803], the NSF CAREER award [2044149], and the Toyota Research Institute (TRI). This article solely reflects the opinions and conclusions of its authors and not ONR, NSF, TRI or any other Toyota entity.}
}

\begin{document}

\maketitle
\thispagestyle{empty}
\pagestyle{empty}

\begin{abstract}
This paper presents an approach for learning motion planners that are accompanied with probabilistic \emph{guarantees} of success on new environments that hold \emph{uniformly} for any disturbance to the robot's dynamics within an admissible set. We achieve this by bringing together tools from generalization theory and robust control. First, we curate a library of motion primitives where the robustness of each primitive is characterized by an over-approximation of the forward reachable set, i.e., a ``funnel". Then, we optimize probably approximately correct (PAC)-Bayes generalization bounds for training our planner to compose these primitives such that the entire funnels respect the problem specification. We demonstrate the ability of our approach to provide strong guarantees on two simulated examples: (i) navigation of an autonomous vehicle under external disturbances on a five-lane highway with multiple vehicles, and (ii) navigation of a drone across an obstacle field in the presence of wind disturbances.
\end{abstract}

\section{Introduction}


Imagine a drone trained to navigate obstacles in an indoor environment, but deployed outdoors where it must contend with previously unseen obstacle environments and wind disturbances. Can we \emph{guarantee} that the learned policy will \emph{generalize} well to such outdoor environments with varying obstacle geometries/placements and exogenous disturbances? Similarly, autonomous vehicles often encounter novel traffic patterns and road conditions, and robotic manipulators encounter objects with novel geometries and friction properties. Learning motion planners that \emph{provably} generalize to scenarios which are statistically dissimilar to those encountered during training is currently an open problem and a crucial limiting factor for the deployment of learning-based approaches in safety-critical applications. In this paper, we provide an approach to learn motion planners with probabilistic guarantees of success in novel environments (e.g., guaranteeing that the drone will remain collision-free with high probability when deployed outdoors with novel obstacle placements) which hold uniformly for \emph{any} disturbance to the robot's dynamics (e.g., wind gusts or parameter mismatch) within an admissible set.


Machine learning approaches are capable of synthesizing motion planners that deal with rich sensory feedback and/or complex scenarios, potentially involving multiple external agents \cite{sunderhauf2018limits,baker2019emergent}.  Recently, approaches for furnishing \emph{generalization guarantees} for learned policies have been proposed \cite{majumdar2020pacbayes,Veer20}; however, these guarantees are intimately coupled with the distribution from which the training environments are drawn. For instance, the generalization guarantees for a drone trained in a particular wind profile are rendered invalid when the wind profile changes. In contrast, approaches from robust control theory leverage the robot's dynamical model to generate policies that lack such fragility; however, the guarantees associated with such techniques are usually limited to low-level control tasks and typically do not extend to systems with rich sensory feedback (e.g., vision) and/or complex scenarios. Evidently, learning and robust control exhibit complementary benefits; however, combining the benefits of both while making guarantees on performance remains challenging. 

\begin{figure}[t]
\centering
\subfigure[]
{
\includegraphics[width=0.23\textwidth]{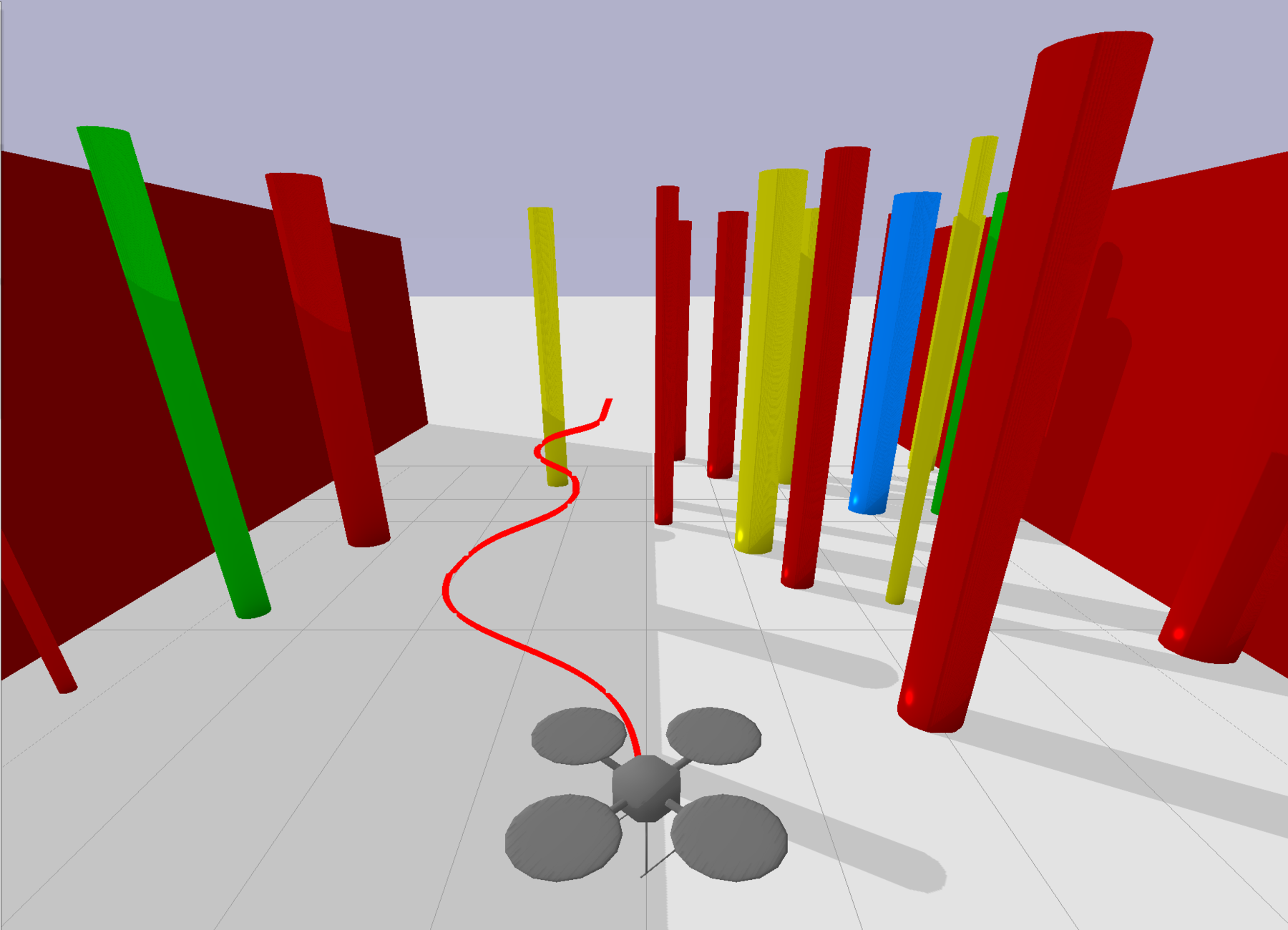}
\label{fig:no-disturb}
}
\hspace{-5mm}
\subfigure[]
{
\includegraphics[width=0.23\textwidth]{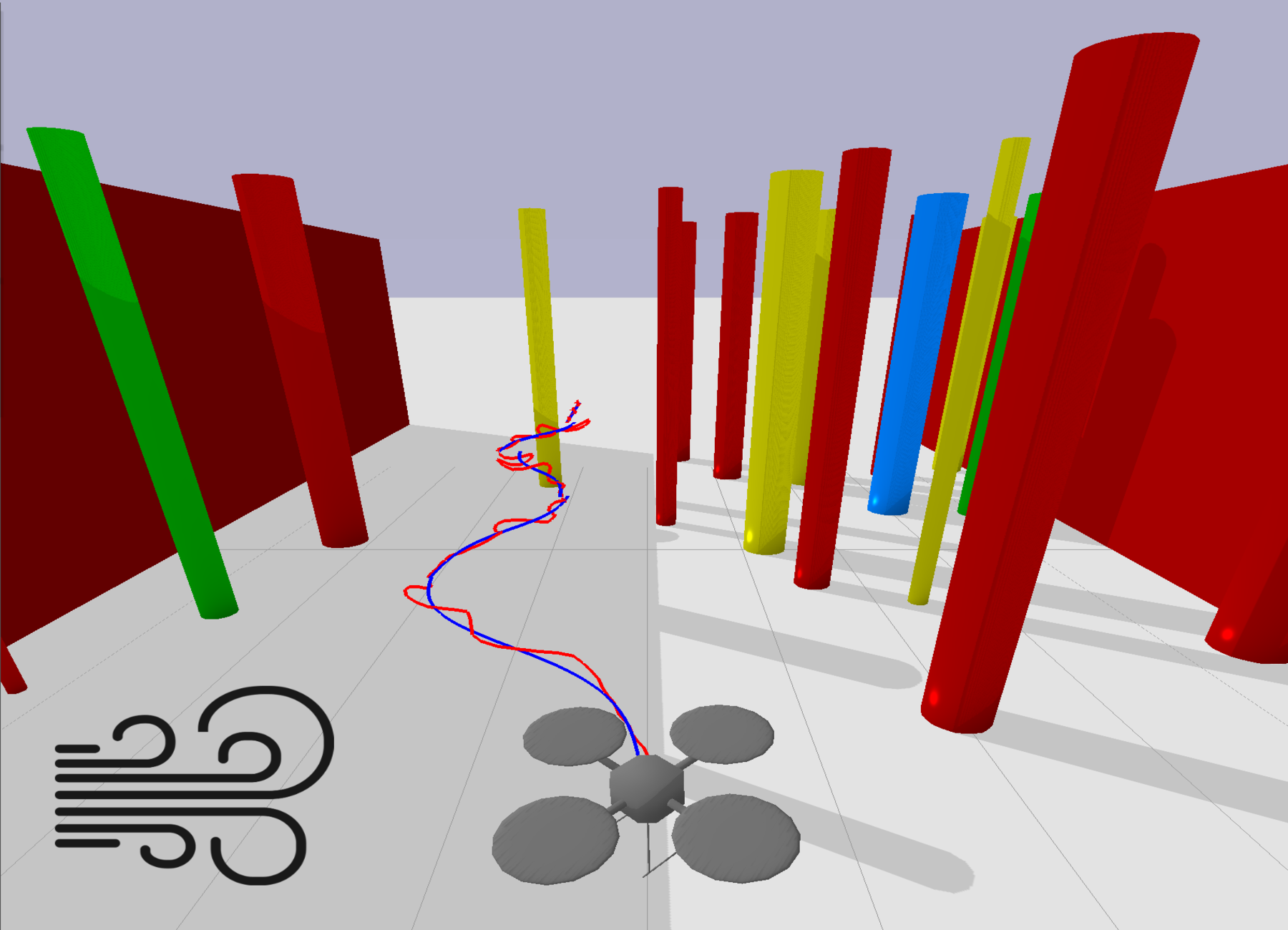}
\label{fig:disturb}
}
\vskip -10pt
\caption{A drone navigating through an environment with obstacles. \textbf{(a)} Motion planner is trained in the absence of disturbances. \textbf{(b)} Motion planner is tested in the presence of external wind disturbances. The blue line represents the nominal (undisturbed) trajectory of the robot and red line represents the trajectory of the robot under disturbances. \label{fig:anchor}
}
\vspace{-5mm}
\end{figure}

\emph{Statement of Contributions:} The main contribution of this paper is a framework for furnishing \emph{probably approximately correct (PAC)} generalization guarantees of success for robots deployed in novel environments that hold \emph{uniformly} for any disturbances to the robot dynamics within an admissible set. We achieve this by bridging learning and robust control via the construction of a library of motion primitives, each member of which is a robust controller accompanied with an over-approximation of the forward reachable set under disturbances --- colloquially known as a ``funnel" \cite{burridge1999sequential,tedrake2010lqr,Majumdar17,mitchell2005time}. Given a set of training environments, we then leverage \emph{PAC-Bayes theory} \cite{McAllester99} in order to learn a policy for switching among the motion primitives such that the entire funnel respects the problem specification. 
This ensures that the PAC-Bayes generalization bounds hold uniformly for any disturbance to the robot dynamics within the admissible set. 
The strength of our theoretical framework and training algorithm is demonstrated on two simulated examples: (a) navigation of an autonomous vehicle under parametric uncertainty on a five-lane highway with multiple vehicles, and (b) navigation of a drone across an obstacle field in the presence of exogenous wind disturbances (Fig. \ref{fig:anchor}).

\section{Related Work}

\textbf{Out-of-distribution (OOD) Generalization.} 
One of the biggest challenges in reinforcement learning (RL) is the brittleness of trained agents to distribution shifts in the environment. Recent studies have developed benchmarks to quantify the generalization performance of RL agents in out-of-distribution environments \cite{cobbe2019quantifying, ahmed2020causalworld, zhang2018study}. Indeed, this problem is particularly relevant to the field of robot learning where policies are often trained in simulation and directly transferred to hardware, resulting in an OOD deployment of the policy due to the mismatch between the simulator and the real world. 
This challenge, known as the sim-to-real gap, is typically addressed by domain randomization methods \cite{tobin2017domain, james2019simtoreal, openai2019solving}. Several works have also focused on learning low-dimensional representations of the world for improved zero-shot domain adaptation \cite{higgins2018darla, dittadi2021representation}. Meta-reinforcement learning frameworks have also been proposed to improve OOD generalization performance \cite{houthooft2018evolved, kirsch2020improving, plappert2018multigoal}. In addition, there has been recent interest in exploiting invariances in the robot's environments for improved OOD generalization \cite{zhang2020learning, sonar2021invariant, agarwal2021contrastive}. However, none of the approaches mentioned above provide guarantees on OOD generalization. An exception to this is the literature on distributionally robust learning \cite{sinha2017certifying} which provides OOD generalization guarantees only for distributions lying within a Wasserstein ball around the training distribution \cite{smirnova2019distributionally,derman2020distributional,ren2021distributionally}. In this paper, we provide generalization guarantees that hold despite \emph{worst-case} disturbances to the robot dynamics.

\textbf{Robust Motion Planning.} Robust motion planning addresses the challenge of safe planning for robots in the presence of uncertainties. 
Probabilistic guarantees on the robot's safety have been presented with sampling-based techniques \cite{janson2015monte, Sun2015HighFrequencyRU}, Gaussian processes \cite{helwa2018provably}, and belief space planning \cite{prentice2009, Platt2012NonGaussianBS, mohammadi14}. Any guarantees provided by these approaches are strongly coupled with the distribution on the robot dynamics. On the other hand, approaches that leverage over-approximations of the reachable states i.e., ``funnels," provide worst-case guarantees that hold uniformly for any uncertainty within a set.
Approaches to estimate the funnels include approximating non-linearities as bounded disturbances \cite{Althoff14}, sums-of-squares programming \cite{parrilo2000structured} for polynomial dynamics \cite{tedrake2010lqr,Majumdar17}, Hamilton-Jacobi-Bellman reachability \cite{lygeros1999controllers,mitchell2005time}, and control contraction metrics \cite{manchester2017control, singh2017}.
Most of these approaches do not directly work with high-dimensional visual feedback and usually make simplifying assumptions, e.g., assumptions on the dynamics of non-ego vehicles on the road or their interactions with the autonomous vehicle. In this paper, we deal with vision and complex interaction scenarios by leveraging the power of learning in combination with robust planning techniques similar to those discussed above.

\textbf{Generalization Guarantees.} Generalization theory in the field of supervised learning studies the ability of a learned function to generalize to novel data \cite{Shalev14}. Generalization is usually quantified by probably approximately correct (PAC) bounds on the expected loss on novel data. PAC generalization bounds have been used in robotics \cite{Karydis15} and controls \cite{Vidyasagar01,Campi19} for providing guarantees on learned models or controllers with low dimensionality.
The PAC-Bayes framework \cite{McAllester99} is a specific family of bounds in generalization theory that have recently been successful in providing generalization bounds for deep neural networks (DNNs) \cite{Dziugaite17, Rivasplate19}.
In our previous work, we developed the PAC-Bayes Control framework \cite{majumdar18, majumdar2020pacbayes} for synthesizing control policies that provably generalize to novel environments. We also provided approaches to furnish generalization bounds for learning vision-based motion planners \cite{Veer20} and for imitation learning \cite{Ren20}. However, none of these papers are able to provide generalization guarantees in the presence of worst-case disturbances to the robot's dynamics. In this paper, we combine the framework proposed in \cite{Veer20} with tools from robust control for computing funnels to provide generalization guarantees of success in novel environments that hold uniformly for any disturbance to the robot dynamics within an admissible set.

\section{Problem Formulation}
\label{sec:prob-form}

We consider robotic systems with continuous-time dynamics. 
For any time $t \geq 0$, let $x(t)\in\mathcal{X}\subseteq\mathbb{R}^n$ be the state of the robot, let $u(t)\in\mathcal{U}\subseteq\mathbb{R}^m$ be the control input, and $w(t)\in\mathbb{R}^d$ be an exogenous disturbance. The dynamics of the system can then be expressed as follows:
\begin{align}\label{eq:dynamics}
    \dot{x}(t) = f(x(t),u(t),w(t)),
\end{align}
where $f:\mathcal{X}\times\mathcal{U}\times\mathbb{R}^d\to\mathbb{R}^n$ is the vector field. We assume knowledge of the parametric form of the dynamics; uncertain parameters can be absorbed in $w$.


We denote the space of the robot's environments by $\mathcal{E}$; ``environment" broadly refers to all effects that influence the robot's performance, excluding disturbances $w$; e.g., motion of other vehicles on the road for a self-driving car or placement and geometry of obstacles for a drone navigation task. We assume access to a set of training environments $S:=\{E_1,E_2, \cdots, E_N\}$ that are drawn from an underlying distribution $\mathcal{D}$ on $\mathcal{E}$. It is worth emphasizing that we \emph{do not} assume any explicit knowledge of $\mathcal{D}$ or $\mathcal{E}$ --- we only have indirect access to these through the finite training dataset $S$. Additionally, we assume that test (i.e., deployment) environments are drawn from the same underlying distribution $\mathcal{D}$.

Let $\mathcal{W}:=\{w:[0,\infty)\to\mathbb{R}^d~|~\|w\|_\infty\leq \gamma\}$ be the space of all disturbance signals (with supremum norm no greater than $\gamma>0$) that affect the robot's dynamics (e.g., wind gust signals). The approach we present in this paper allows us to train the motion planner \emph{in the absence} of disturbances, i.e., $w\equiv 0$, but still provide guarantees for any disturbance in $\mathcal{W}$ at deployment. 
In particular, we provide generalization guarantees on the success of the robot in novel environments drawn from the distribution $\mathcal{D}$ on $\mathcal{E}$ which hold uniformly over the set of disturbances to the robot dynamics $\mathcal{W}$. 


We assume that the robot is equipped with proprioceptive sensors $h:\mathcal{X}\to\mathcal{Y}$ (e.g., IMUs) and exteroceptive sensors $g:\mathcal{X}\times\mathcal{E}\to\mathcal{O}$ (e.g., RGB-D cameras); note that our sensors do not provide feedback for disturbances. We define a motion primitive as a low-level controller $\Gamma:[0,\infty)\times\mathcal{Y}\to\mathcal{U}$ which tracks a trajectory in the state space $\mathcal{X}$ of the robot's dynamics. Let $\mathcal{J}$ be a compact index set and $\mathcal{L}\coloneqq\{\Gamma_{j}\colon [0,\infty) \times\mathcal{Y}\to\mathcal{U}\mid j\in\mathcal{J}\}$ be a library of motion primitives which includes a diverse range of motions for the robot. We aim to learn control policies $\pi:\mathcal{O}\to\mathcal{L}$ which decide the motion primitive in the library $\mathcal{L}$ to implement given observations of the environment. Let the rollout of the robot's state trajectory while executing a policy $\pi$ from an initial state $x_0\in\mathcal{X}$, in an environment $E\in\mathcal{E}$, and with disturbances $w\in\mathcal{W}$ be denoted by $r_\pi: [0,\infty)\times\mathcal{X}\times\mathcal{W}\to\mathcal{X}$. We define the cost of deploying a policy $\pi$ in an environment $E$ over a given time horizon $T$ as $c(\pi,E,w)\in[0,1]$. As an example, the cost $c(\pi,E,w)$ for the drone navigation task can be assigned as 0 if the rollout $r_\pi$ intersects with the obstacles or 1 otherwise. 

Our goal is to learn a distribution $P$ over the policy space $\Pi$ that minimizes the expected cost across \emph{novel} environments drawn from $\mathcal{D}$ under worst-case disturbances $w$ belonging to the set $\mathcal{W}$. Thus, we want to solve the following:
\begin{align}
    C^{*} & \coloneqq  \inf_{P \in \mathcal{P}} \sup_{w \in \mathcal{W}} \mathop{\mathbb{E}}_{E\sim\mathcal{D}} \mathop{\mathbb{E}}_{\pi\sim P} [c(\pi,E, w)] \tag{$\mathcal{OPT}$}\label{eq:OPT}
\end{align}
where $\mathcal{P}$ is the space of probability distributions over $\Pi$. 

\section{Approach: PAC-Bayes Funnel Composition}

In this section we present our approach for leveraging PAC-Bayes theory and motion planning with funnel libraries in order to provide guarantees on generalization to new environments that hold uniformly for any disturbance within the admissible set. 

\subsection{Funnel Libraries}\label{subsec:funnel}
The closed-loop dynamics of $\eqref{eq:dynamics}$ for a particular motion primitive $\Gamma_{j}$ can be expressed as:
\begin{align}\label{eq:cl_dyn}
    \dot{x}(t) = f_{j}(x(t),w(t)).
\end{align}
Given an initial condition set $X_j \subseteq \mathcal{X} \subseteq \mathbb{R}^n$, a funnel $F_{j}$ is an over-approximation of the forward reachable set of states that the solutions of \eqref{eq:cl_dyn} are guaranteed to stay within as long as the disturbances $w$ lie within the admissible set $\mathcal{W}$ and the initial state of the system lies within $X_j$:
\begin{align}
    x(0) \in X_j \implies x(t) \in F_j(t), ~\forall t\in[0,T_j], ~\forall w\in\mathcal{W}.
\end{align}

\begin{figure}[t]
\centering
\includegraphics[trim={1cm 3cm 1cm 0cm},clip,width=0.33\textwidth]{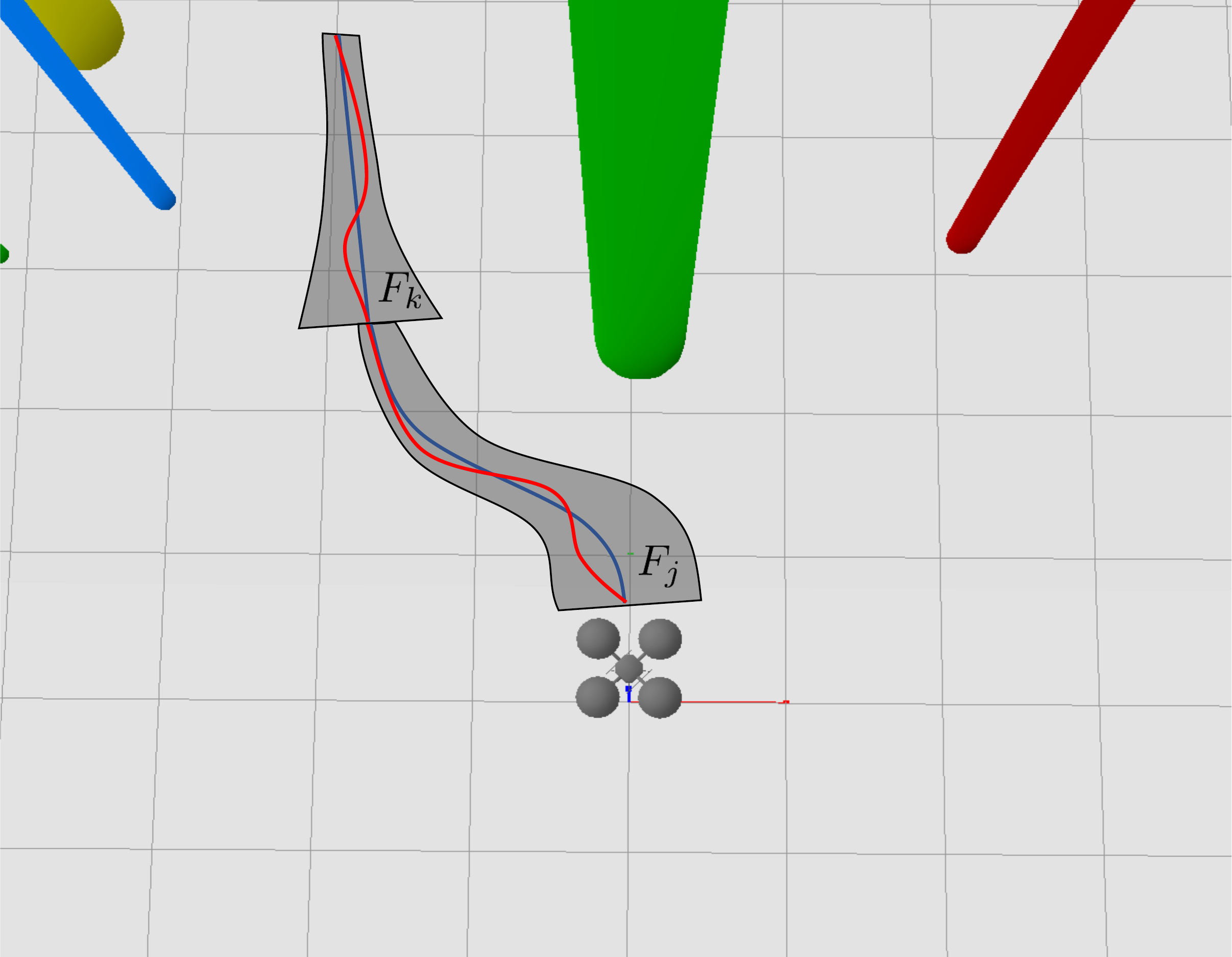}
\caption{Illustration of the composability of funnels $F_j$ and $F_k$. The nominal trajectory in the absence of disturbances is depicted in blue while the trajectory under disturbances is depicted in red.\label{fig:funnel-comp}
}
\vspace{-5mm}
\end{figure}

A funnel library $\mathcal{F} \coloneqq\{F_{j}\colon [0, T_j] \to 2^\mathcal{X}\mid j\in\mathcal{J}\}$ is the collection of the funnels for motion primitives in $\mathcal{L}$; $2^\mathcal{X}$ is the powerset of $\mathcal{X}$. A funnel $F_k$ is considered to be composable with $F_j$ if the inlet $F_k(0)$ of $F_k$ lies within the outlet $F_j(T_j)$ of $F_j$, i.e., $F_j(T_j)\subseteq F_k(0)=X_k$. Satisfaction of the composability criterion ensures that if we execute the primitive $j$ and then $k$, the rollout of \eqref{eq:dynamics} will stay within the set defined by the union of the two funnels; see Fig.~\ref{fig:funnel-comp} for an illustration. For the sake of brevity, the exposition on funnel compositions is kept terse here; see \cite{Majumdar17} for further details.

The policy $\pi:\mathcal{O}\to\mathcal{L}$ is responsible for selecting a motion primitive based on the robot's exteroceptive sensor observations and can be executed in a receding-horizon manner in an environment $E$ (i.e., selecting a primitive, executing the corresponding policy, selecting another primitive, etc.). In this work, we will \emph{learn} a policy that generates a score for each motion primitive and then executes the primitive with the highest score among the subset of composable primitives to \emph{ensure the composability of each funnel}. This results in a sequence of primitive executions $\{j_k\}_{k=1}^K$ in an environment $E$ and a corresponding sequence of funnel compositions $F_\pi:=\{F_{j_k}\}_{k=1}^K$ where each consecutive pair of funnels is composable.

We define $C(\pi,E)$ as the cost associated with the states in the funnel sequence $F_\pi$ rather than the rollout $r_\pi$. For example, in the drone navigation task, the cost associated with the funnel $C$ can assign 0 if the \emph{entire funnel sequence} $F_\pi$ does not intersect with any obstacles and 1 if it does. The funnel-sequence cost $C$ then upper bounds the rollout cost $c$ as demonstrated in the following remark. 
%
\begin{remark}\label{rem:funnel-cost-bound}
Let $w\in\mathcal{W}$ be arbitrary and let the robot's state be initialized in the inlet of the first funnel in $F_\pi$. Since each successive funnel in the sequence is composable with the preceding one, the rollout $r_\pi$ for the policy $\pi$ in environment $E$ is then guaranteed to stay within the union of the funnels in the sequence $F_\pi$. Therefore, the cost for the funnel sequence will upper bound that of the rollout $r_\pi$:
\begin{align}
    c(\pi,E,w) \leq C(\pi,E), ~ \forall w\in\mathcal{W} \enspace.
\end{align}
\end{remark}


A crucial consequence of Remark~\ref{rem:funnel-cost-bound} is that the optimal value of \eqref{eq:OPT} can be bounded from above as follows:
\begin{align}
    C^{*} & \coloneqq \inf_{P \in \mathcal{P}} \sup_{w \in \mathcal{W}} \mathop{\mathbb{E}}_{E\sim\mathcal{D}} \mathop{\mathbb{E}}_{\pi\sim P} [c(\pi,E, w)] \nonumber \\
    & \leq \inf_{P \in \mathcal{P}} \mathop{\mathbb{E}}_{E\sim\mathcal{D}} \mathop{\mathbb{E}}_{\pi\sim P} [C(\pi,E)] =: \inf_{P \in \mathcal{P}} C_{\mathcal{D}}(P) \label{eq:OPT_1}
\end{align}
where $C_{\mathcal{D}}(P)$ is the policy distribution $P$'s expected funnel-sequence cost across novel environments drawn from $\mathcal{D}$. Leveraging \eqref{eq:OPT_1}, we address \eqref{eq:OPT} by minimizing its upper bound $C_\mathcal{D}$ using PAC-Bayes generalization bounds.


\subsection{PAC-Bayes Generalization Guarantees}\label{subsec:PAC-Bayes}
As mentioned in Section~\ref{sec:prob-form}, we do not assume an explicit characterization of $\mathcal{D}$ and only assume a finite dataset of training environments; it is thus infeasible to directly minimize $C_\mathcal{D}(P)$ in \eqref{eq:OPT_1}. We address this challenge via PAC-Bayes generalization theory \cite{McAllester99}, which provides an upper bound on $C_\mathcal{D}(P)$ in terms of the cost on the training environments and a regularizer. In particular, we use the vision-based PAC-Bayes control framework introduced in \cite{Veer20}.


Let $\Pi := \{\pi_{\theta} \mid \theta \in \RR^q \}$ denote the space of policies $\pi$ parameterized by $\theta$; e.g., $\Pi$ can be the space of neural networks with a fixed architecture with $\theta$ as the weights. For a particular choice of the ``posterior" policy distribution $P$ on $\Pi$ and a dataset $S \coloneqq \{E_{1}, E_{2}, \dotsm,E_{N} \}$ of $N$ training environments drawn i.i.d from $\mathcal{D}$, we define the \emph{empirical cost} as the expected cost across the environments in $S$:
\begin{align}
    C_{S}(P) \coloneqq \frac{1}{N} \sum_{E \in S} \mathop{\mathbb{E}}_{\theta \sim P} [C(\pi_{\theta}, E)].
    \label{eq:emp_cost}
\end{align}

Let $P_0$ represent a ``prior" distribution over $\Pi$ which is specified before observing the training data $S$. The PAC-Bayes Control theorem then provides an upper bound on the true expected cost $C_\mathcal{D}(P)$ which holds with high probability.

\begin{theorem}[adapted from \cite{Veer20}]
For any $\delta\in(0,1)$ and posterior $P$, with probability at least $1-\delta$ over sampled environments $S\sim \mathcal{D}^N$, the following inequality holds:
\begin{align}
~C_{\mathcal{D}}(P) & \leq C_{PAC}(P,P_0) \nonumber \\
&  := \big(\sqrt{C_S(P) + R(P,P_0)} + \sqrt{R(P,P_0)}\big)^2 , \label{eq:quad-pac-bound}
\end{align}
where
$R(P,P_0)$ is a regularization term defined as:
\begin{equation}\label{eq:R}
R(P,P_0):=\frac{\textrm{KL}(P||P_0) + \log\big(\frac{2\sqrt{N}}{\delta}\big) }{2N} \enspace.
\end{equation} 
\end{theorem}
This result provides an upper bound on the true cost $C_\mathcal{D}(P)$ that holds with probability at least $1-\delta$. Therefore, we can indirectly tackle \eqref{eq:OPT} by minimizing \eqref{eq:quad-pac-bound}. Additionally, the final PAC bound $C_{PAC}(P,P_0)$ provides a certificate of the motion planner's generalization performance on novel environments drawn from $\mathcal{D}$ and for \emph{any} disturbance $w\in\mathcal{W}$. The ability to provide such worst-case guarantees across disturbances $w$ distinguishes the approach in this paper from our prior work in \cite{Veer20}. 

\section{Training and Deployment} 
In this section we discuss the training pipeline for obtaining the posterior distribution $P$ which is deployed in novel environments. It is worth pointing out that the tightness of the PAC-Bayes bound is heavily reliant on the choice of the prior $P_0$. We thus split our training dataset into two parts $\hat{S}$ and $S$ and use $\hat{S}$ for training a strong prior $P_0$. Utilizing the trained prior, we minimize the upper bound of \eqref{eq:quad-pac-bound} for the funnel-sequence cost $C$ to solve \eqref{eq:OPT}. In the interest of space we only provide a sketch of the training pipeline and point to \cite{Veer20} (which shares a similar algorithmic approach for training) for further details. 

\textbf{Training a Prior with ES.}
We train the prior by minimizing the empirical cost on $\hat{S}$ using a class of blackbox optimization methods called Evolutionary Strategies (ES) \cite{Wierstra2014}. ES allows parallelization of rollouts and thus facilitates the effective exploitation of cloud computing resources. Let $P_0$ belong to the family of multivariate Gaussian distributions $\mathcal{N}(\mu, \Sigma)$ with mean $\mu\in\mathbb{R}^q$ and a diagonal covariance $\Sigma\in\mathbb{R}^{q\times q}$ (which we represent by the element-wise square-root of its diagonal vector $\sigma\in\mathbb{R}^q$). 
Adapting \eqref{eq:emp_cost} for $\hat{S}$, we can express the gradient of the empirical cost with respect to $\psi:=(\mu,\sigma)\in\mathbb{R}^{2q}$ as follows:
\begin{equation}\label{eq:emp-grad-psi}
\nabla_\psi C_{\hat{S}}(P_0) = \frac{1}{\hat{N}} \sum_{E \in \hat{S}} \nabla_\psi\underset{\theta \sim P_0}{\mathbb{E}} [C(\pi_\theta; E)] \enspace.
\end{equation}
A Monte-Carlo estimate of this gradient is passed to the Adam optimizer \cite{Kingma14} to update $\psi$ for minimizing $C_{\hat{S}}$.

\textbf{Training a PAC-Bayes Policy.} To obtain the posterior, we minimize the PAC-Bayes upper bound \eqref{eq:quad-pac-bound}. If the probability distributions on the policy space are discrete, this minimization problem can be transformed to an efficiently-solvable convex program \cite{Veer20}. Therefore, we first restrict the continuous policy space $\Pi$ to the finite policy space $\Tilde{\Pi}:=\{\pi_{\theta_i}~|~\theta_i\sim P_0,~i=1,\cdots,m\}$ which includes a finite sampling of policies from $P_0$. The prior for the PAC-Bayes minimization with the finite-policy space $\Tilde{\Pi}$ is then chosen as the uniform distribution $p_0$. The discrete posterior distribution is denoted by $p$. To facilitate the convex program, we first compute $C\in\mathbb{R}^m$ which is the policy-wise cost vector; each entry of $C$ is the average cost of deploying policy $\pi_i \in \tilde{\Pi}$ on the environments $\{E_i\}_{i=1}^N$ in the training dataset $S$. The empirical cost $C_S(p)$ can then be expressed as a linear function of $p$, i.e., $Cp$. The minimization of the PAC-Bayes bound \eqref{eq:quad-pac-bound} can be written as:
\begin{align}
\min_{p\in\mathbb{R}^m} \quad & \big(\sqrt{Cp + R(p,p_0)} + \sqrt{R(p,p_0)}\big)^2 \label{eq:REP}\\
\textrm{s.t.} \quad & \sum_{i=1}^m p_i = 1, 0\leq p_i \leq 1. \nonumber
\end{align}
The transformation of \eqref{eq:REP} into a convex program and the exact algorithm for solving it are provided in \cite[Sec.~4]{Veer20}.

\textbf{Deployment of the PAC-Bayes Policy.} At deployment in any new test environment, we execute a single deterministic policy sampled from the finite-policy space $\Tilde{\Pi}$ according to the PAC-Bayes posterior $p$ obtained by solving \eqref{eq:REP}. The presence of disturbances in a rollout might affect the distribution on the observations received by the policy, potentially leading the policy to execute a different primitive than what it would have in the absence of disturbances; e.g., the depth maps at the end of the first primitive execution in the absence of disturbances (Fig.~\ref{fig:no-disturb}) and in the presence of disturbances (Fig.~\ref{fig:disturb}) may not be identical. However, this does not pose a major challenge to our approach, primarily because the funnels of our motion primitives ensure that despite disturbances the robot trajectories stay close to the nominal trajectories in the absence of disturbances; see Fig.~\ref{fig:funnels}. Such boundedness despite disturbances can usually be achieved by choosing low-level controllers $\Gamma_j$ with sufficiently high gains. Hence, the terminal state of each primitive under disturbances is in close proximity to the terminal state in the absence of disturbances, ensuring close correspondence of the observations. Indeed, our results --- discussed in the next section --- validate the ability of our framework to furnish generalization guarantees that hold under disturbances.


\section{Results} In this section we demonstrate our theoretical framework on two simulated examples: (a) navigation of an autonomous vehicle under disturbances due to parametric uncertainty on a five-lane highway with multiple other vehicles and (b) navigation of a UAV across a static obstacle field in the presence of wind disturbances. Through these examples we demonstrate the ability of our framework to furnish strong generalization guarantees of success that hold uniformly for any disturbance to the dynamics within an admissible set.

\begin{figure}
\centering
\subfigure[]
{
\includegraphics[width=0.22\textwidth]{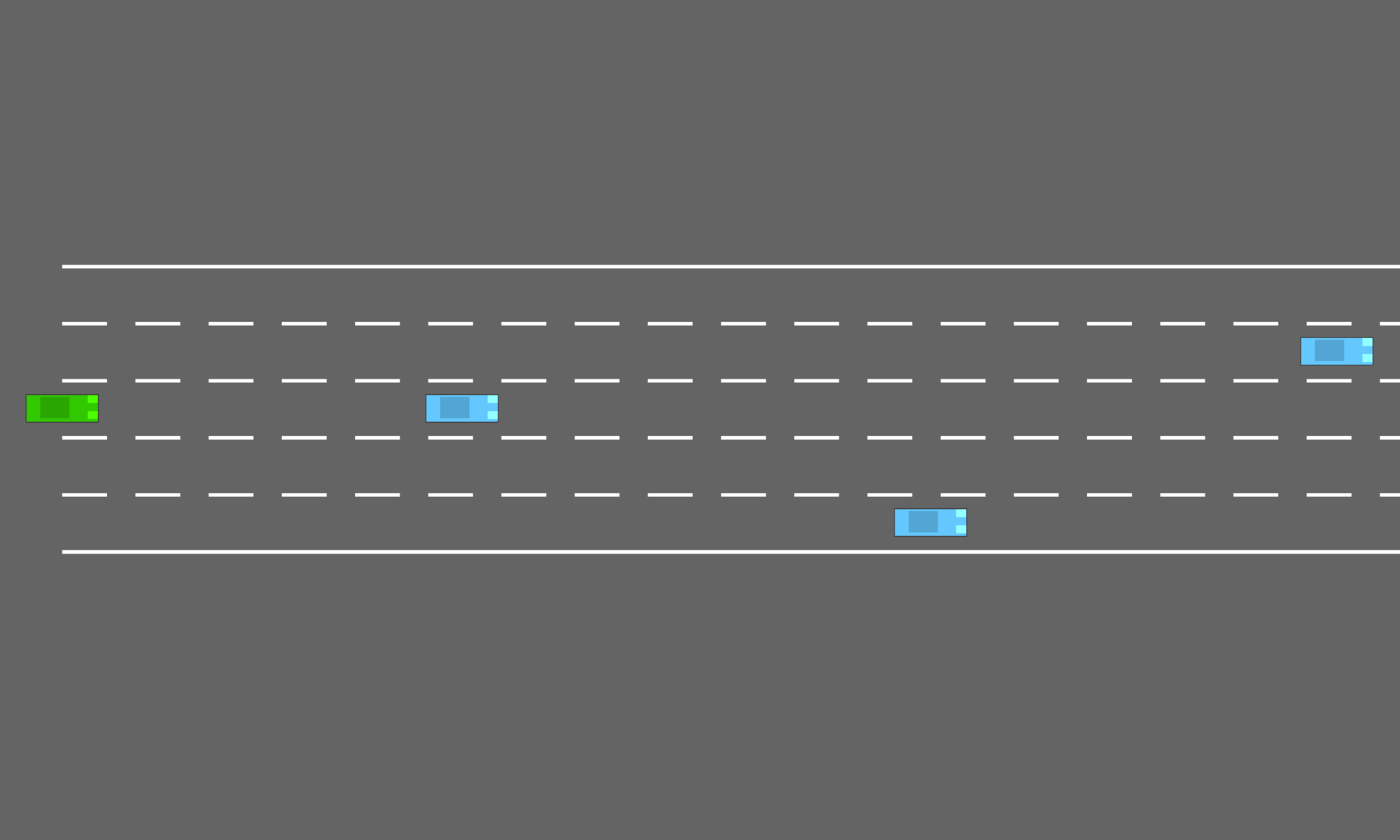}
\label{fig:env-car}
}
\centering
\hspace{-2mm}
\subfigure[]
{
\includegraphics[width=0.22\textwidth]{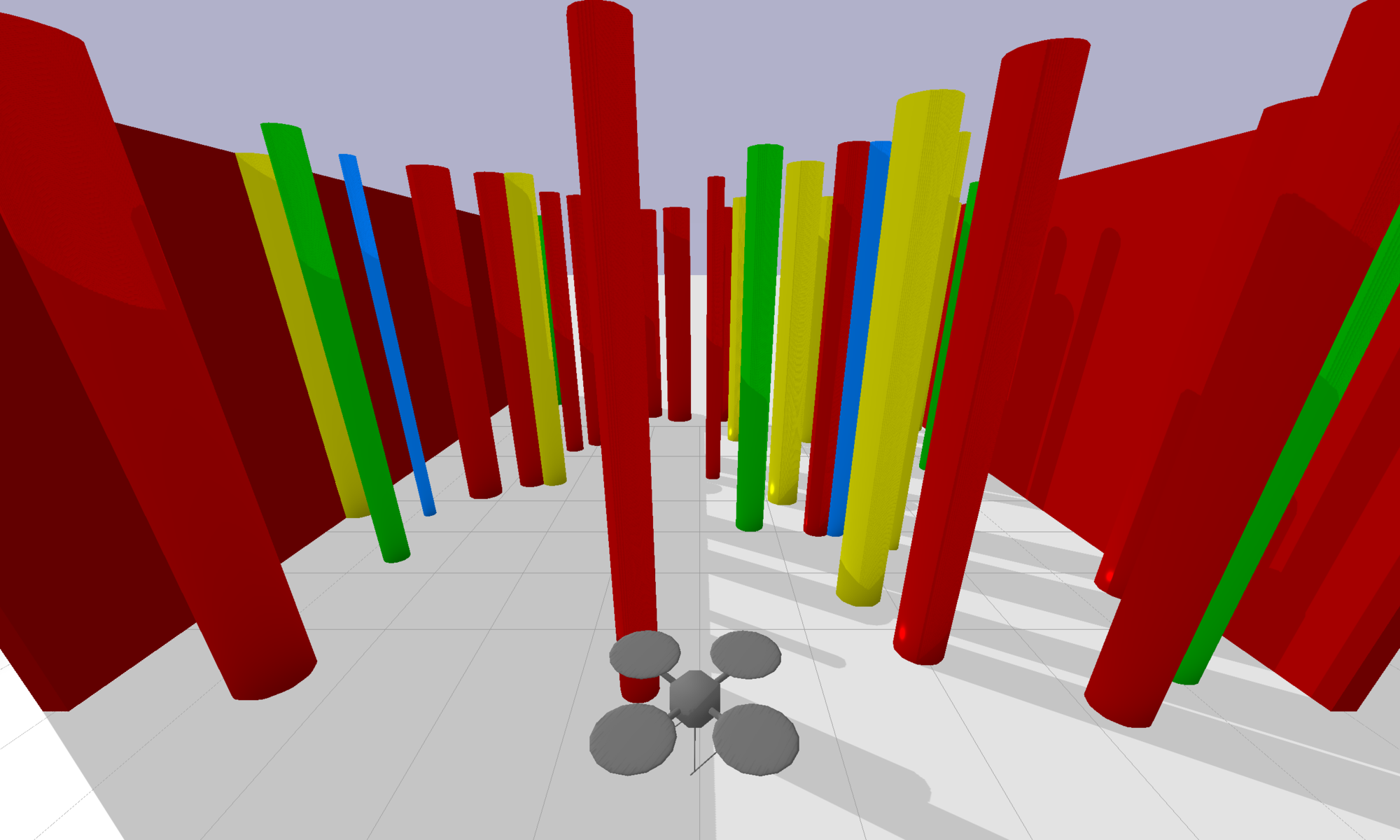}
\label{fig:env-quad}
}
\vskip -10pt
\caption{Examples for robust learning-based planning. \textbf{(a)} Simulation of a vehicle under external disturbances navigating on a road with multiple other vehicles. \textbf{(b)} Simulation of a drone navigating across a static obstacle field in the presence of force disturbances.\label{fig:envs}
}
\vspace{-5mm}
\end{figure}

\subsection{Autonomous Vehicle Navigation}
In this example we train an autonomous vehicle to navigate a five-lane highway with other moving vehicles (Fig. \ref{fig:env-car}).

\textbf{Environment.} We adapt the ``\texttt{highway-v0}" gym environment \cite{highway-env} for this example. In our environment, the forward velocity component of the ego vehicle is constant at $10~\textrm{m/s}$; the sideways component can vary to permit lane changes. All other vehicles on the road have time-varying forward velocities drawn from a uniform distribution on $[0~\textrm{m/s}, 9~\textrm{m/s}]$ and they always stay in their lane. Effectively, our ego vehicle must weave through the traffic to get ahead. We assume that the ego vehicle has exteroceptive sensors $g:\mathcal{X}\times\mathcal{E}\to\mathcal{O}$ that observe the relative position of the five nearest vehicles. Even though our sensor provides the exact relative position of other vehicles, directly planning motions for the future based on this information might yield poor results due to the lack of information of the other cars' velocities and intentions; for instance, other cars might speed up or slow down, rendering the motion plan infeasible. The uncertain intention and velocity of other cars, therefore, makes it convenient to design the planner via learning. 

\textbf{Funnels.} The ego vehicle is modelled by the kinematic bicycle model with external disturbances:
\begin{align}\nonumber
    & \dot{x} = v \cos(\theta) + w_{x};~ \dot{y} = v \sin(\theta) + w_{y};~ \dot{\theta} = v \frac{\tan(\zeta)}{L} + w_{\theta}
\end{align}
where $L$ is the vehicle's length, $v$ is the car's speed in the current heading direction, $\theta$ is the heading angle, and $\zeta$ is the steering angle which is also the control input of the car. External disturbances $w_{x}$, $w_{y}$ and $w_{\theta}$ are drawn from uniform distributions on $[-0.5~\textrm{m/s}, 0.5~\textrm{m/s}]$, $[-1~\textrm{m/s}, 1~\textrm{m/s}]$ and $[-0.25~\textrm{rad/s}, 0.25~\textrm{rad/s}]$, respectively.

\begin{figure}[t]
\centering
\subfigure[]
{
\includegraphics[width=0.22\textwidth]{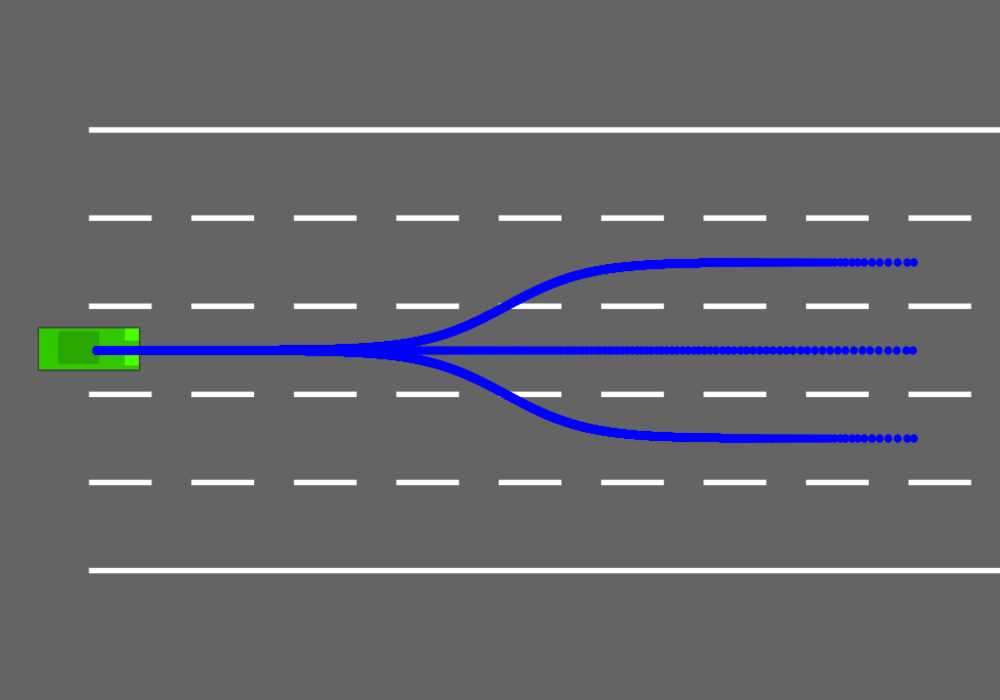}
\label{fig:car_prims}
}
\centering
\hspace{-2mm}
\subfigure[]
{
\includegraphics[width=0.22\textwidth]{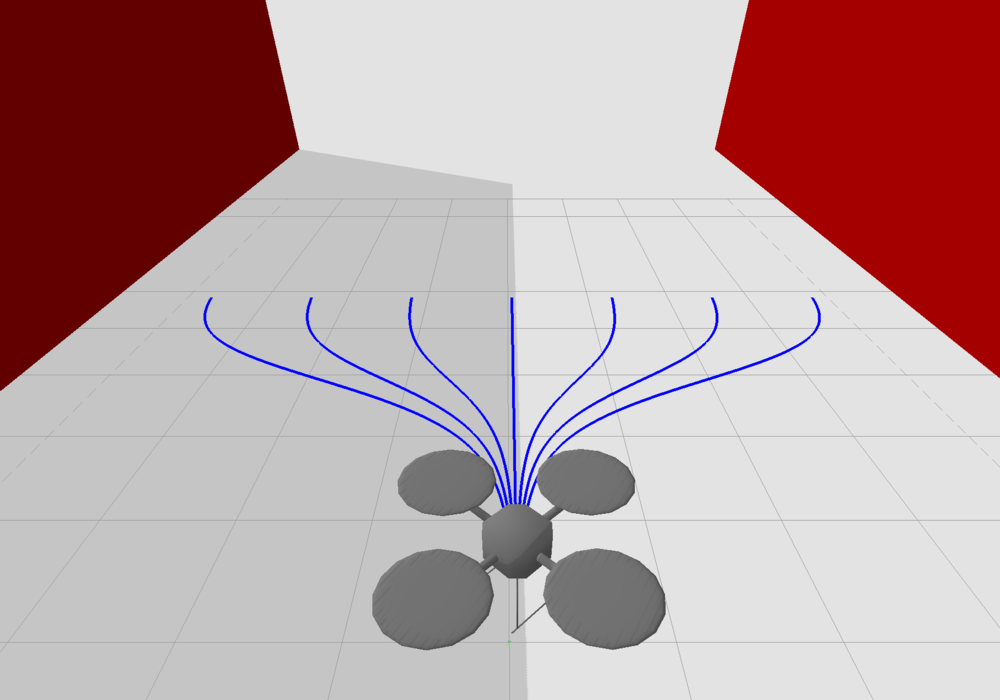}
\label{fig:quad_prims}
}
\vskip -10pt
\caption{Motion primitives for the two examples. \textbf{(a)} Motion primitive library of the ego vehicle. \textbf{(b)} Motion primitive library of the drone.\label{fig:prims}
}
\vspace{-5mm}
\end{figure}

Our library consists of three motion primitives: (i) change to left lane, (ii) keep current lane, and (iii) change to right lane (Fig. \ref{fig:car_prims}). The motion primitive trajectories are generated by connecting the initial position $(x_{0}, y_{0})$ and the final desired position $(x_{0} + \Delta x, y_{0} + \Delta y)$ by a smooth sigmoidal trajectory. Differential flatness of the kinematic bicycle model is used to recover the full states from the desired motion primitive trajectories. We implement a PD controller for tracking the desired position and orientation during the execution of a motion primitive using the model parameters and control gains in \cite{wheeler}. Note that the uncertainties in the system cause the vehicle to deviate from its nominal motion primitive trajectory and the controller tries to correct for these deviations.

\begin{figure}[t]
\centering
\subfigure[]
{
\includegraphics[width=0.22\textwidth]{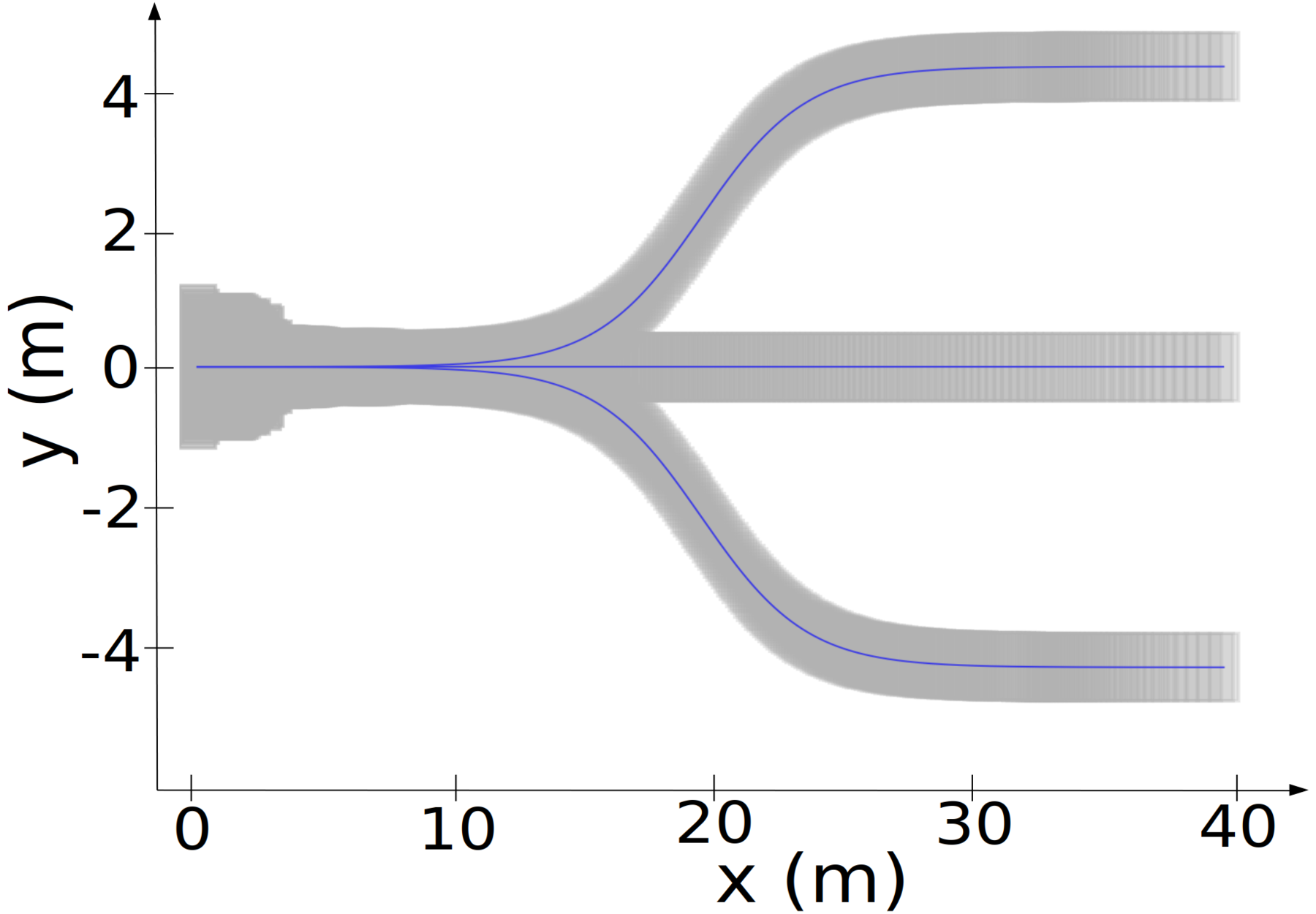}
\label{fig:car_funnels}
}
\centering
\hspace{-2mm}
\subfigure[]
{
\includegraphics[width=0.22\textwidth]{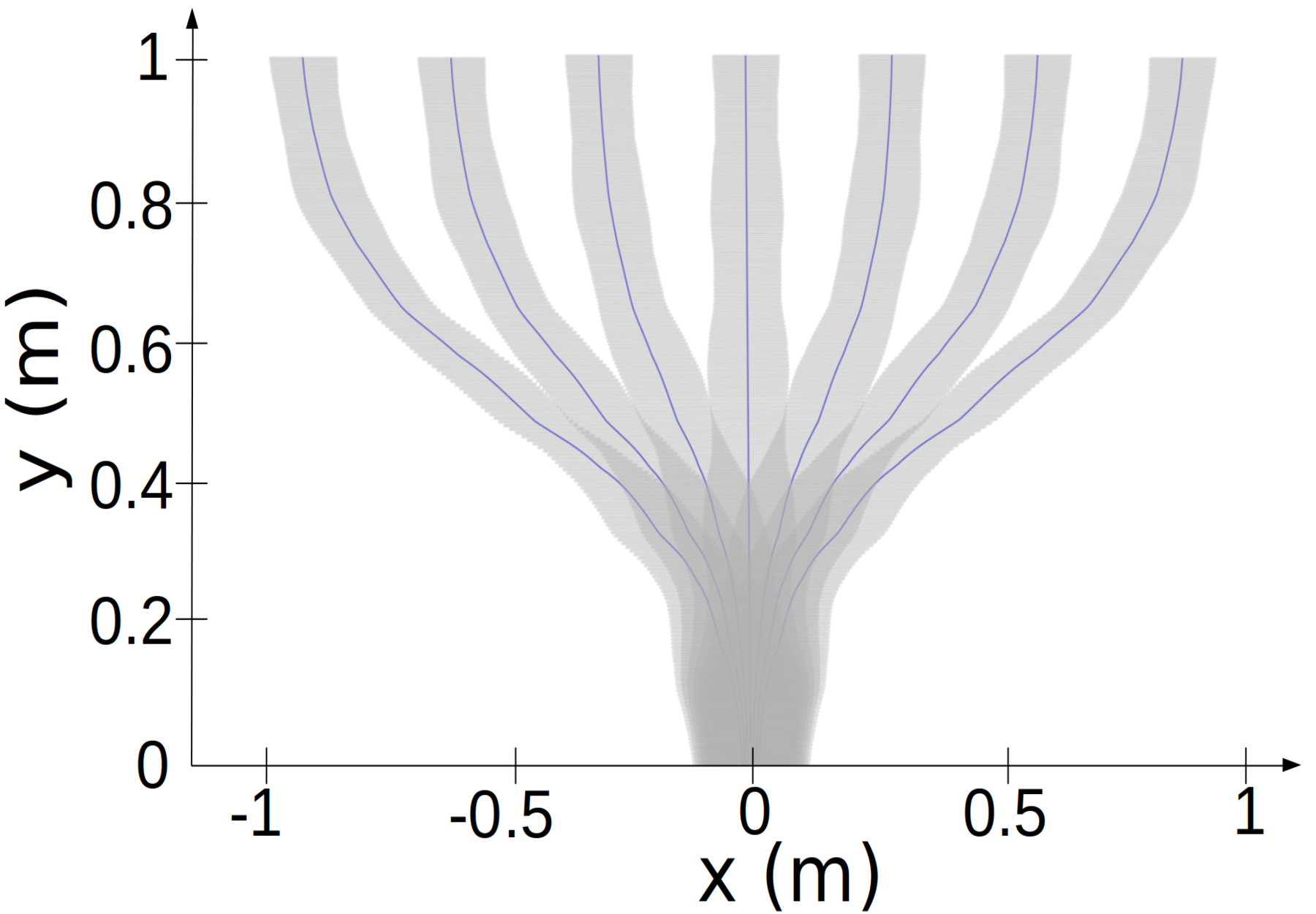}
\label{fig:quad_funnels}
}
\vskip -10pt
\caption{Funnel libraries for the two examples projected onto the $X-Y$ plane; underlying motion primitives are shown with blue lines. \textbf{(a)} Funnel library of the autonomous vehicle. \textbf{(b)} Funnel library of the drone.\label{fig:funnels}
}
\vspace{-5mm}
\end{figure}

We perform offline reachability analysis computations using the JuliaReach toolbox \cite{JuliaReach19} to generate the funnels around the nominal motion primitive trajectories (Fig. \ref{fig:car_funnels}). JuliaReach's TMJets algorithm is utilized and the reachable sets are over-approximated using hyper-rectangles. TMJets is  a  Taylor  model-based algorithm  that  uses  Taylor  polynomials  with  guaranteed  error  bounds  to approximate the nonlinear dynamics.  We use Taylor model approximations of order 3 since higher order approximations were computationally challenging. The computation time for each funnel was approximately 2 minutes, running on a 4.5 GHz laptop with 16 GB memory and 12 cores. We ensure that the inlet of any funnel $F_k$ lies within the outlet of any other funnel $F_j$, thus leading to a library of three funnels that are all sequentially composable with each other.

\textbf{Policy.} Our DNN-based control policy $\pi : \mathcal{O} \to \mathcal{L}$ maps an observation to a score vector ($\in \mathbb{R}^3$) and selects the motion primitive with the highest score. The input to the neural network are the relative positions of the five nearest vehicles. The policy network consists of 4 fully connected layers and has 1043 parameters.

\textbf{Training.} We choose the cost $1 - k/K$ where $k$ is the number of motion primitives successfully executed in a rollout before colliding with another vehicle or the boundaries of the highway and $K$ is the maximum possible primitive executions in a rollout; in our example $K = 10$. The training time for the prior is $\sim 30$ hours running on a server with 50 CPUs. The PAC-Bayes optimization is performed on the same machine with a run-time of $\sim 2000$ sec.

\textbf{Results.} The PAC-Bayes results are detailed in Table~\ref{tab:car-results}. We set $\delta = 0.01$ to have PAC-Bayes bounds hold with probability 0.99. We compute the bound by drawing 40 i.i.d. policies from the trained prior $P_{0}$ and computing the cost for each policy in 4000 training environments drawn i.i.d. from $\mathcal{D}$. This computation provides us a PAC-Bayes generalization bound of $16.23\%$, which can be verbally interpreted as: \emph{on average the ego vehicle will successfully navigate through at least 83.77\% ($100\% - 16.23\%$) of the novel environments}. Finally, to empirically estimate the true expected cost incurred by the trained posterior, we perform exhaustive simulations on novel environments with and without external disturbances, yielding $13.77\%$ and $14.44\%$, respectively. As expected, the cost estimate in the presence of disturbances is upper bounded by our PAC-Bayes bound of $16.23\%$. 

For comparison, we train a new motion planner with the same set of environments and training settings with nominal primitive trajectories \emph{instead of funnels} and in the absence of disturbances; the results are detailed in Table~\ref{tab:car-results}. The PAC-Bayes bound of $16.66\%$ is violated in the presence of disturbances and the true cost (under disturbances) $17.38\%$ is worse compared to the policy trained with funnels ($14.44\%$). This demonstrates the importance of the funnels for the generalization bounds to hold under disturbances and the benefits of incorporating them in the training pipeline.
\vspace{-3mm}
\begin{table}[H]
  \caption{PAC-Bayes Results for Vehicle Navigation \label{tab:car-results}}
  \vspace{-2mm}
  \centering
  \begin{tabular}{|c|c|cc|}
  	\hline
    Funnels & PAC-Bayes Bound & \multicolumn{2}{c|}{True Cost (Estimate)} \\
    & $C_{PAC}$ & No Dist. & Dist. \\
    \hline
    Yes & 16.23\% & 13.77\% & 14.44\% \\
    No & 16.66\% & 15.78 \% & 17.38\% \\
    \hline
  \end{tabular}
  \vspace{-5mm}
\end{table}

\subsection{Vision-based Drone Navigation}
In this section we train a drone to navigate an obstacle-dense course in the presence of external disturbances using depth maps from an \emph{onboard} vision sensor (Fig. \ref{fig:env-quad}).

\textbf{Environment.} The simulation environment, similar to the environment in \cite{Veer20}, was built using PyBullet \cite{pybullet}. The environment has 50 cylindrical obstacles. The (unknown) distribution $\mathcal{D}$ over environments is chosen by drawing obstacle radii and locations from a uniform distribution on $[5~\textrm{cm}, 30~\textrm{cm}]$ and $[-5~\textrm{m}, 5~\textrm{m}] \times [0~\textrm{m}, 14~\textrm{m}]$, respectively. The drone’s goal is to move forward using its motion primitives while avoiding any collision with obstacles in the environment. Bounded uncertainties are present in the form of force disturbances to the robot dynamics.

\textbf{Funnels.} The drone is modeled using the full quadrotor dynamics with 12-dimensional state space. We introduce external wind disturbances by adding bounded disturbance terms $w_{wind} \in [-0.1~\textrm{m}/\textrm{s}^2, 0.1~\textrm{m}/\textrm{s}^2]\times [-0.025~\textrm{Nm}, 0.025~\textrm{Nm}]$ to the linear accleration and body moment in the control input.
These disturbances cause the robot to deviate from its nominal trajectory. We implement a nonlinear tracking feedback controller \cite{Mellinger11} that can correct for these deviations and track the nominal trajectory in the 12-dimensional state space. We perform offline reachability analysis computations using the JuliaReach toolbox \cite{JuliaReach19} to generate funnels around the nominal motion primitive trajectories (Fig. \ref{fig:quad_prims}) using the same algorithm and parameters as in the autonomous vehicle example. We ensure that the inlet of any funnel $F_k$ lies within the outlet of any other funnel $F_j$, thus leading to a library of seven funnels that are all sequentially composable with each other (Fig. \ref{fig:quad_funnels}). The computation time for each funnel was  $\sim 5$ minutes, running on a 4.5 GHz laptop with 16 GB memory and 12 cores.

\textbf{Policy.} Our DNN-based control policy $\pi : \mathcal{O} \to \mathcal{L}$ maps a depth map observation ($50 \times 50$) to a score vector $(\in\mathbb{R}^7)$ and selects the motion primitive with the highest score. We model our policy architecture based on the vision-based UAV navigation policy presented in \cite{Veer20}.

\textbf{Training.} We choose the cost $1 - \frac{k}{K}$ where $k$ is the number of motion primitives successfully executed before colliding with an obstacle and $K$ is the total possible primitive executions; in our example $K = 14$. The training time for the prior is $\sim 62$ hours running on a server with 4 NVIDIA RTX 2080 GPUs and 32 CPUs. PAC-Bayes optimization is performed on the same server with $\sim 4000$ sec run-time.

\textbf{Results.} The PAC-Bayes results are detailed in Table~\ref{tab:drone-results}. We set $\delta = 0.01$ to have PAC-Bayes bounds hold with a probability 0.99. We compute the PAC-Bayes bound by drawing 40 i.i.d. policies from the trained prior $P_{0}$ and computing the cost for each policy in 4000 training environments drawn i.i.d. from $\mathcal{D}$. This computation provides us a PAC-Bayes generalization bound of $19.7\%$, which can be verbally interpreted as: \emph{on average the UAV will successfully navigate through at least 80.3\% ($100\% - 19.7\%$) of the novel environments}. Performing exhaustive simulations on novel environments to empirically estimate the true expected cost achieved by the trained posterior with and without disturbances results in $15.7\%$ and $19.0\%$, respectively; see Table~\ref{tab:drone-results} for a summary of the results.

\vspace{-3mm}
\begin{table}[H]
  \caption{PAC-Bayes Results for UAV Navigation \label{tab:drone-results}}
  \vspace{-2mm}
  \centering
  \begin{tabular}{|c|cc|}
  	\hline
    PAC-Bayes Bound & \multicolumn{2}{c|}{True Cost (Estimate)} \\
    $C_{PAC}$ & No Dist. & Dist. \\
    \hline
    19.7\% & 15.7\% & 19.0\% \\
    \hline
  \end{tabular}
  \vspace{-2mm}
\end{table}

\section{Conclusions and Future Work}

In this paper, we presented an approach to train motion planners accompanied with generalization guarantees of success in new environments that hold uniformly for any dynamic disturbances within an admissible set. We achieved this by characterizing funnels for each motion primitive and then training our planner to compose primitives --- such that the entire funnels satisfy the problem specifications --- by optimizing a PAC-Bayes generalization bound. The efficacy of our approach to provide strong performance bounds despite disturbances is demonstrated by our examples.

\textbf{Future Work.} This work gives rise to numerous exciting future directions. 
To ensure the composability of the funnels in this paper, we proposed the implicit approach of executing the primitive with the highest policy-generated score from among the subset of composable primitives. We believe that the performance of our approach would greatly benefit by explicitly embedding logic specifications in the policy while learning \cite{leung2020back,aksaray2016q}. Embedding such structure would facilitate faster training and potentially better generalization.
%
Another exciting future direction is to augment our approach with methods from safe reinforcement learning \cite{brunke2021safe} to provide generalization guarantees on the success of safety-critical systems while ensuring safety during the training process itself (in contrast to only providing guarantees for deployment, as we do here). Finally, we are working towards hardware experiments of our approach on a drone navigating an obstacle field under wind disturbances. 






\bibliography{citations}
\bibliographystyle{IEEEtran}

\end{document}